\documentclass[sn-mathphys-num]{sn-jnl}

\usepackage{graphicx}%
\usepackage{multirow}%
\usepackage{amsmath,amssymb,amsfonts}%
\usepackage{amsthm}%
\usepackage{mathrsfs}%
\usepackage[title]{appendix}%
\usepackage{xcolor}%
\usepackage{textcomp}%
\usepackage{manyfoot}%
\usepackage{booktabs}%
\usepackage{algorithm}%
\usepackage{algorithmicx}%
\usepackage{algpseudocode}%
\usepackage{listings}%

\usepackage{amsmath,amsfonts,bm}

\def\eqref#1{equation~\ref{#1}}

\def\1{\bm{1}}

\def\vm{{\bm{m}}}

\def\vp{{\bm{p}}}

\def\mA{{\bm{A}}}

\def\mH{{\bm{H}}}
\def\mI{{\bm{I}}}

\def\mW{{\bm{W}}}
\def\mX{{\bm{X}}}

\DeclareMathAlphabet{\mathsfit}{\encodingdefault}{\sfdefault}{m}{sl}
\SetMathAlphabet{\mathsfit}{bold}{\encodingdefault}{\sfdefault}{bx}{n}

\newcommand{\R}{\mathbb{R}}

\usepackage{multicol}
\usepackage[T1]{fontenc}
\usepackage{pifont}
\usepackage{ulem}
\usepackage[capitalize,noabbrev]{cleveref}
\newcommand{\cmark}{{\color{blue}\ding{51}}}%
\newcommand{\xmark}{{\color{red}\ding{55}}}%
\newcommand{\tblidx}[1]{{\small \texttt{[#1]}}}

\theoremstyle{thmstyleone}%
\theoremstyle{thmstyletwo}%

\theoremstyle{thmstylethree}%

\begin{document}

\title[Article Title]{A Generalist Cross-Domain Molecular Learning Framework for Structure-Based Drug Discovery}


\author[1,2]{\fnm{Yiheng} \sur{Zhu}}
\equalcont{These authors contributed equally to this work.}

\author[1]{\fnm{Mingyang} \sur{Li}}
\equalcont{These authors contributed equally to this work.}

\author[1]{\fnm{Junlong} \sur{Liu}}

\author[1]{\fnm{Kun} \sur{Fu}}

\author[3]{\fnm{Jiansheng} \sur{Wu}}

\author[1]{\fnm{Qiuyi} \sur{Li}}

\author[1,2]{\fnm{Mingze} \sur{Yin}}

\author[1]{\fnm{Jieping} \sur{Ye}}

\author*[4,5]{\fnm{Jian} \sur{Wu}}\email{wujian2000@zju.edu.cn}

\author*[1]{\fnm{Zheng} \sur{Wang}}\email{wz388779@alibaba-inc.com}

\affil[1]{\orgname{Alibaba Cloud Computing}, \orgaddress{\city{Beijing}, \postcode{100012}, \country{China}}}

\affil[2]{\orgdiv{College of Computer Science and Technology}, \orgname{Zhejiang University}, \orgaddress{\city{Hangzhou}, \postcode{310058}, \state{Zhejiang}, \country{China}}}

\affil[3]{\orgdiv{School of Computer Science}, \orgname{Nanjing University of Posts and Telecommunications}, \orgaddress{\city{Nanjing}, \postcode{210023}, \state{Jiangsu}, \country{China}}}

\affil[4]{\orgdiv{State Key Laboratory of Transvascular Implantation Devices of The Second Affiliated Hospital}, \orgname{Zhejiang University School of Medicine}, \orgaddress{\city{Hangzhou}, \postcode{310058}, \state{Zhejiang}, \country{China}}}

\affil[5]{\orgdiv{School of Public Health}, \orgname{Zhejiang University}, \orgaddress{\city{Hangzhou}, \postcode{310058}, \state{Zhejiang}, \country{China}}}


\abstract{
Structure-based drug discovery (SBDD) is a systematic scientific process that develops new drugs by leveraging the detailed physical structure of the target protein. Recent advancements in pre-trained models for biomolecules have demonstrated remarkable success across various biochemical applications, including drug discovery and protein engineering. However, in most approaches, the pre-trained models primarily focus on the characteristics of either small molecules or proteins, without delving into their binding interactions which are essential cross-domain relationships pivotal to SBDD. To fill this gap, we propose a general-purpose foundation model named BIT (an abbreviation for Biomolecular Interaction Transformer), which is capable of encoding a range of biochemical entities, including small molecules, proteins, and protein-ligand complexes, as well as various data formats, encompassing both 2D and 3D structures. Specifically, we introduce Mixture-of-Domain-Experts (MoDE) to handle the biomolecules from diverse biochemical domains and Mixture-of-Structure-Experts (MoSE) to capture positional dependencies in the molecular structures. The proposed mixture-of-experts approach enables BIT to achieve both deep fusion and domain-specific encoding, effectively capturing fine-grained molecular interactions within protein-ligand complexes. Then, we perform cross-domain pre-training on the shared Transformer backbone via several unified self-supervised denoising tasks. Experimental results on various benchmarks demonstrate that BIT achieves exceptional performance in downstream tasks, including binding affinity prediction, structure-based virtual screening, and molecular property prediction. Furthermore, we develop a BIT-driven virtual screening pipeline that has identified two hit compounds with compelling inhibitory activity against the GluN1/GluN3A N-methyl-D-aspartate (NMDA) receptor, as validated by wet-lab assays. The code and pre-trained models will be made publicly available.}

\keywords{structure-based drug discovery, molecular representation learning, molecular interaction, multimodal learning}



\maketitle

\section{Introduction}\label{sec1}
Structure-based drug discovery (SBDD) is a systematic scientific strategy that aims to identify potential drug candidates by thoroughly analyzing the physical structures of target proteins, including analyzing the intricate structure of the target, understanding its function, and designing molecules capable of interacting with the target in a specific and favorable manner to regulate its activity. To complement the labor-intensive traditional methods, geometric deep learning algorithms~\citep{atz2021geometric} have recently been proposed to improve the efficiency and performance of various stages of the SBDD process~\citep{zhang2023systematic}, including binding site identification~\citep{sverrisson2021fast}, binding affinity prediction~\citep{li2021structure}, virtual screening~\citep{torng2019graph}, \textit{de novo} molecule design~\citep{luo20213d}, etc. 

Over the last few years, the self-supervised pre-training of foundation models has revolutionized the fields of natural language processing~\cite{devlin2018bert,brown2020language} and computer vision~\citep{chen2020simple,he2022masked}. Inspired by this unprecedented success, significant efforts have been dedicated to molecular pre-training, aiming to exploit the vast potential inherent in the extensive corpus of unlabeled molecules, particularly small molecules and proteins~\citep{xia2023molebert, ferruz2022controllable}. Fine-tuning pre-trained models can significantly enhance performance across various biochemical downstream tasks, such as molecular property prediction~\citep{rong2020self} and protein structure prediction~\citep{lin2023evolutionary}. However, most existing approaches are specialized for a single data domain, focusing exclusively on either small molecules or proteins. This specialization limits the ability of pre-trained models to capture molecular interactions across different biochemical domains.

Protein-ligand interactions are crucial in orchestrating biological processes at the molecular level~\citep{tomasi1994molecular}. Understanding the fundamental principles that underlie these interactions is crucial in scientific fields, as it facilitates a broad range of downstream applications, especially in the context of SBDD~\citep{anderson2003process,vamathevan2019applications}. To model molecular interactions and advance the process of SBDD, the prevailing practice usually involves either training task-specific models from scratch~\citep{li2021structure, luo20213d, corso2023diffdock} or using a simple interaction module that combines pre-trained molecular and protein encoders in a task-specific manner~\citep{zhou2023unimol}. However, these models are often prone to overfitting, especially when the assay-labeled data is scarce. Additionally, these task-specific designs make it challenging to leverage pre-training effectively, as they may not fully capture intricate interaction patterns. Therefore, unlocking the full potential of pre-training to enhance interaction-related downstream tasks remains a significant challenge.

\begin{figure*}[t]
    \centering
    \includegraphics[width=0.72\linewidth]{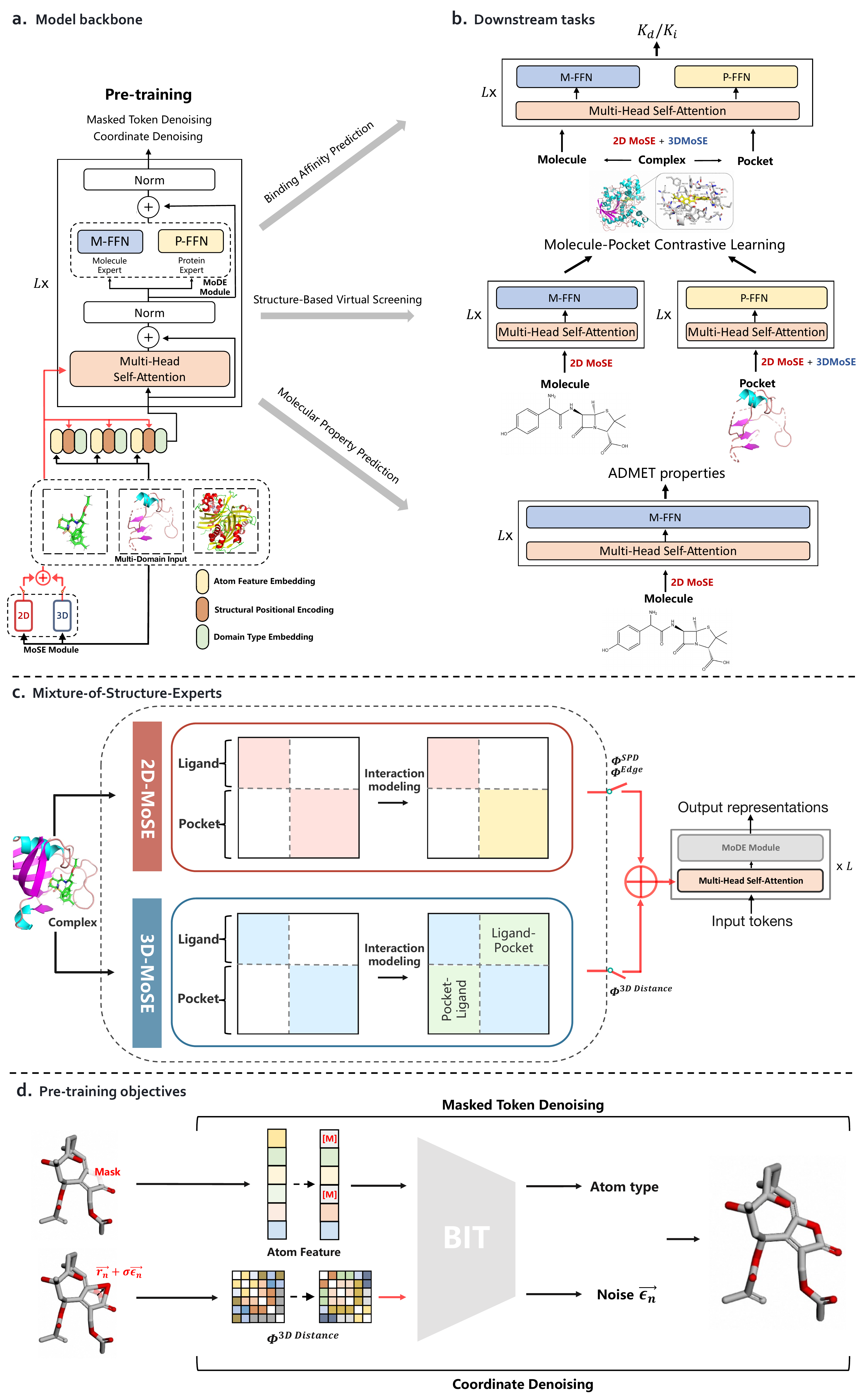}
    \caption{BIT overview. a) We employ a general-purpose Transformer model as the backbone network to carry out masked token denoising and coordinate denoising tasks on protein-ligand complex data, as well as on unbound small molecule and pocket datasets. Additionally, we introduce Mixture-of-Domain-Experts (MoDE) and Mixture-of-Structure-Experts (MoSE) within specific modules to capture multi-domain specificity and inter-domain relationships. b) BIT services as a foundation model with diverse functionalities, including a fusion encoder for binding affinity prediction, a dual encoder for virtual screening, or a molecule encoder for molecular property prediction. c) We introduce the MoSE, which utilizes specialized pairwise bias expert networks tailored for different domains.
    In the 2D-MoSE, we transition from a shared pairwise bias expert for small molecules and pockets to an independent pairwise bias expert for each entity. Conversely, in the 3D-MoSE, we maintain the shared expert for each entity, while introducing independent bias experts specifically for the protein-ligand interaction modeling. Distinct parameters within these networks are denoted by varying colors. d) We propose unified corrupt-then-denoise objectives (i.e., coordinate denoising and masked token denoising) applicable to various domain data.}
    \label{fig:framework}
\end{figure*}

Recent efforts have focused on pre-training models explicitly designed to capture cross-domain dependencies between protein pockets and ligands, as demonstrated by CoSP~\citep{gao2023co} and DrugCLIP~\citep{gao2023drugclip}. These methods differentiate the two domains as independent signals and adopt CLIP~\citep{radford2021learning} to learn a shared embedding space where bindable pockets and ligands are pulled closer. However, simply aligning the embeddings of bindable molecules does not capture the nuanced interaction details. Consequently, both CoSP and DrugCLIP fall short in effectively addressing complex protein-ligand binding tasks, such as binding affinity prediction, which rely heavily on such detailed information. Inspired by the remarkable achievements in multimodal learning~\citep{radford2021learning, li2022blip, bao2022vlmo, wang2023image, xu2023multimodal} (\cref{app_related_work}), we believe that \textit{it is promising to harness essential information from diverse biochemical domains and build more powerful pre-trained models that support both domain-specific encoding and cross-domain interactions.}

To refine and optimize the SBDD process, we present a general-purpose model called the \textbf{B}iomolecular \textbf{I}nteraction \textbf{T}ransformer (\textbf{BIT}) following the \textit{protein-ligand pre-training} paradigm, as depicted in~\cref{fig:framework}. BIT encodes molecules across a wide range of biochemical domains, including small molecules, proteins, and protein-ligand complexes, as well as diverse data formats, encompassing both 2D and 3D structures, all within a unified Transformer backbone. The backbone is constructed upon Transformer-M~\citep{luo2023one}, a model renowned for its flexibility and effectiveness in handling both 2D and 3D structural data. We further enhance it to capture both multi-domain specificity and inter-domain relationships by incorporating \textit{Mixture-of-Domain-Experts} (MoDE) and \textit{Mixture-of-Structure-Experts} (MoSE) approaches. In each Transformer block, MoDE replaces the feed-forward network with two distinct domain experts: the molecule expert and the protein expert. Concurrently, MoSE introduces separate domain-specific structural channels to bias attention, yet preserves a shared self-attention module across domains to facilitate alignment between different domains. In BIT, each input atom token is routed to its respective domain/structure expert, allowing the BIT to function as a fusion encoder to model molecular interactions in protein-ligand complexes, or as a dual encoder to independently encode small molecules and proteins.

To learn more precise cross-domain representations, we pre-train BIT on protein-ligand complexes with 3D cocrystal structures~\citep{wei2023biolip2}, as well as on large-scale unbound small molecules and pockets with 3D equilibrium structures. This process is conducted within a unified framework utilizing denoising tasks for both continuous atom coordinates and categorical atom types. We demonstrate BIT's superior performance through extensive experiments across various downstream tasks, including both protein-ligand interaction and molecular learning. As a fusion encoder in binding affinity prediction, BIT consistently outperforms specialized baselines by a decent margin. Additionally, when used as a dual encoder in virtual screening, BIT still achieves state-of-the-art performance while offering significantly faster inference speed. Furthermore, BIT outperforms related state-of-the-art pre-trained models in numerous molecular property prediction tasks. We also conduct ablation studies to validate the effectiveness of the key design choices in pre-training. Ultimately, by integrating BIT into a virtual screening pipeline, we successfully identify two hit compounds with notable inhibitory activity against the GluN1/GluN3A N-methyl-D-aspartate (NMDA) receptor, with the most effective compound showing a half maximal inhibitory concentration ($\text{IC}_{50}$) of 2.67 \textmu M.

The main contributions of this work are summarized as follows:
\begin{itemize}
\item We present BIT, a general-purpose foundation model designed to encode a range of biochemical entities, including small molecules, proteins, and protein-ligand complexes, across various data formats, encompassing both 2D and 3D structures, all by a unified Transformer backbone.
\item We introduce a unified pre-training strategy for BIT on protein-ligand complexes with 3D cocrystal structures, alongside large-scale unbound small molecules and protein pockets with 3D equilibrium structures, to learn more precise cross-domain molecular representations.
\item Experiments confirm that BIT achieves exceptional performance in downstream protein-ligand binding and molecular learning tasks. Further wet-lab experiments underscore BIT's broad applicability and significant potential in SBDD.
\end{itemize}

\section{Results}\label{sec2}
In this section, we begin with a brief overview of the BIT framework. We then provide a comprehensive evaluation of BIT using well-established public benchmarks, covering both protein-ligand binding tasks and molecular learning tasks. Subsequently, we perform an ablation study to investigate the impact of different model components and training strategies on performance. Finally, we integrate BIT into a virtual screening pipeline to identify compounds targeting GluN1/GluN3A NMDA receptors. Further details are available in Methods (\cref{sec:method}).

\subsection{Overview of BIT}
As illustrated in~\cref{fig:framework}, BIT is a general-purpose pre-trained model designed to encode molecules across various biochemical domains, including small molecules, proteins, and protein-ligand complexes, in different data formats, including 2D and 3D structures. BIT can be fine-tuned as a fusion encoder to model intricate molecular interactions within protein-ligand complexes for precise binding affinity prediction, a dual encoder to enable efficient virtual screening, or a unimodal encoder for modeling small molecules (\cref{fig:framework}b). To achieve this purpose, we treat diverse molecules at the single atom level and introduce a shared Transformer backbone for unified modeling, a unified pre-training strategy to learn more precise cross-domain representations, and a flexible fine-tuning strategy for task-specific adaptation.

In biochemical applications, data are collected in the form of molecules represented at different levels of granularity, such as atoms, residues, and nucleobases. However, all molecules can be uniformly represented as sets of atoms held together by attractive or repulsive forces. To more effectively transfer atom-level knowledge across different domains, we propose to share atom embeddings and incorporate domain embeddings to distinguish between small molecules and proteins. Besides, for protein-ligand complexes with cocrystal structures, we identify the binding pocket as the protein atoms located within a minimum distance of 5 {\AA} from the ligand~\citep{muegge1999general}. Then we input the extracted pocket-ligand complex into BIT to learn contextualized representations. It is noteworthy that we only use the binding pocket as the model input rather than the entire protein primarily for the following two reasons: (1) the binding pocket is the paramount region of protein-ligand interaction, experiencing the most significant spatial alterations during the binding process and providing sufficient insight into molecular interactions; (2) the binding pocket contains significantly fewer atoms than the entire protein, leading to lower computational costs and faster training speeds.

The backbone network of BIT, shown in~\cref{fig:framework}a, is built upon Transformer-M~\citep{luo2023one}, a model renowned for its versatility and effectiveness in processing both 2D and 3D molecule data. Briefly, Transformer-M introduces two separate channels to encode 2D and 3D structural information, which are then integrated as bias terms in the multi-head self-attention (MSA) module. To further encode molecules across biochemical domains and facilitate the learning of cross-domain molecular representations enriched with molecular interaction knowledge, we propose two extensions to Transformer-M. Firstly, we introduce the \textit{Mixture-of-Domain-Experts} (MoDE) to effectively handle the biomolecules from various biochemical domain. As shown in~\cref{fig:framework}a, each Transformer block in BIT consists of a shared MSA module and two feed-forward networks (FFNs), presenting domain experts, namely the molecule expert and the protein expert. In contrast to conventional mixture-of-experts layer~\citep{shazeer2017outrageously,fedus2022switch}, which routes input tokens by a trainable gating network, we directly assign an expert to process each atom token based on its molecule data domain. Secondly, we introduce the \textit{Mixture-of-Structure-Experts} (MoSE), which utilizes specialized pairwise bias expert networks tailored for different domains. This mechanism is necessitated by the significant disparities in distributions of molecular structures across biochemical domains, particularly between small molecules and protein pockets. As depicted in~\cref{fig:framework}c, MoSE is delicately designed based on the observation and analysis of 2D and 3D structures from various domains. For the 2D pairwise bias, distinct bias experts are employed for different domains. For the 3D pairwise bias, one set of parameters is used to learn intra-molecular distances, while another set is dedicated to learning inter-molecular distances. These enhancements, collectively referred to as MoD(S)E, allow BIT to enable both deep fusion and domain-specific encoding, as well as to capture fine-grained inter-molecular interactions within protein-ligand complexes featuring 3D cocrystal structures.

We pre-train BIT on protein-ligand complex data, in addition to unbound small molecule and pocket datasets (\cref{sec:pretrain}). We use the Q-BioLiP database~\citep{wei2023biolip2} as the complex corpus. To prevent potential overfitting to a limited portion of the chemical space represented by the Q-BioLiP dataset, we additionally incorporate the PCQM4Mv2 dataset~\citep{nakata2017pubchemqc}, which has been widely used for 3D molecular pre-training~\citep{zaidi2022pre, wang2023automated}, and extract potential pockets on proteins from the Protein Data Bank~\citep{berman2000protein}. To ensure the scalability of the pre-training process, we propose unified corrupt-then-denoise objectives applicable to various domain data (\cref{fig:framework}d). During pre-training, we randomly corrupt the continuous atom coordinates and the categorical atom types of single-domain molecules (i.e., unbound small molecules and pockets) and ligands from protein-ligand complexes, and guide BIT to restore the original states. The \textit{coordinate denoising task}, interpreted as learning an approximate molecular force field from equilibrium structures~\citep{zaidi2022pre}, aims to derive meaningful representations that elucidate the inter-atomic interactions within a molecular structure. Besides, the \textit{masked token denoising task} seeks to capture the fundamental physicochemical properties of molecules or complexes by modeling the dependencies among their atoms. More detailed formulations can be found in~\cref{sec:objectives}.

Thanks to MoD(S)E, BIT effectively decouples the encoding process across various domains, thereby serving as a general-purpose foundation model. As illustrated in~\cref{fig:framework}b, BIT can be further fine-tuned to function as a fusion encoder for protein-ligand binding affinity prediction, a dual encoder for structure-based virtual screening, or a unimodal encoder for molecular property prediction, each configuration being specifically tailored to meet the requirements of the respective downstream tasks.

\subsection{Protein-Ligand Binding Affinity Prediction}
To demonstrate the effectiveness of BIT, we first evaluate it on the protein-ligand binding affinity prediction task. In this task, the pre-trained model serves as a fusion encoder and is fine-tuned to predict binding affinities $pK_a$ (or $-\log K_d$, $-\log K_i$) for protein-ligand complexes with known 3D structures. Following previous studies~\citep{li2021structure}, we perform experiments using two public datasets: (\expandafter{\romannumeral1}) PDBbind v2016~\citep{wang2004pdbbind,wang2005pdbbind} which is a standard benchmark for assessing the performance of models designed to predict binding affinities. (\expandafter{\romannumeral2}) CSAR-HiQ dataset~\citep{dunbar2011csar} which is an additional benchmark resource, commonly employed as an external dataset to further evaluate the generalization ability of models trained on the PDBbind dataset. We evaluate the prediction performance using Pearson's correlation coefficient (R), Mean Absolute Error (MAE), Root-Mean Squared Error (RMSE), and Standard Deviation (SD)~\citep{su2018comparative}. We present the details of baselines and experiment settings in~\cref{app_exp_pdbbind}.

As presented in Table~\ref{tab:affinity}, BIT consistently outperforms pre-training baselines and other approaches tailored for binding affinity prediction across all evaluation metrics, demonstrating the effectiveness of BIT in capturing intricate fine-grained molecular interactions present in complexes. On the PDBbind core set, all pre-training methods achieve superior performance compared to other sophisticated methods that forego pre-training, implying that it is promising to acquire essential interaction knowledge through pre-training. Moreover, it is noteworthy that BIT exhibits exceptional performance on the CSAR-HiQ dataset. Such an observation indicates that the proposed pre-training strategy has endowed our model with a robust capacity for generalization.

\subsection{Structure-based virtual screening}
\label{sec:vs}
Structure-based virtual screening of potential drug-like molecules against a protein target of interest, as outlined by~\citet{lionta2014structure}, is a critical goal in SBDD. The objective of this task is to identify the molecules that exhibit the highest probability of binding to protein pockets with established 3D structures. We perform experiments using two public datasets: (\expandafter{\romannumeral1}) DUD-E dataset~\citep{mysinger2012directory} which is one of the most popular virtual screening benchmarks. (\expandafter{\romannumeral2}) LIT-PCBA dataset~\citep{tran2020-lit-pcba}, which is a much more challenging virtual screening benchmark, proposed to address the biased data problem faced by other benchmarks. We provide results in terms of the AUC-ROC, ROC enrichment (RE) scores, and Enrichment Factor (EF). The formal definition can be found in~\cref{app_vs}.

Since most of the protein-ligand pairs of interest do not have experimentally solved cocrystal structures, conventional affinity prediction models that rely on this information must be complemented with molecular docking software, such as AutoDock~\citep{trott2010autodock}. However, this integration often leads to significant computational expenses, particularly in large-scale virtual screening tasks. By framing virtual screening as a pocket-to-ligand retrieval task, BIT can be adopted as a dual encoder. We encode 3D protein pockets and 2D molecular graphs separately to obtain their representations in a shared subspace and compute their similarity scores by the dot product. During fine-tuning, BIT is optimized using the contrastive loss function InfoNCE~\citep{oord2018representation}, with 64 randomly sampled decoys per active compound. We present the details of baselines and experiment settings in~\cref{app_vs}.

As presented in Table~\ref{tab:dud} and Table~\ref{tab:lit}, BIT achieves superior performance compared to the baselines, with notably higher RE and EF scores which suggest its impressive ability to prioritize the identification of hit compounds. Besides, BIT attains a high degree of screening efficiency without compromising learning precision, since it does not necessitate the joint encoding of every possible pocket-ligand pair and can retain pre-computed representations of both pockets and ligands. In our empirical analysis, we managed to screen 1B molecules from an ultra-large-scale screening library (e.g., ZINC~\citep{irwin2005zinc} and Enamine REAL~\citep{grygorenko2020generating}) in just under two days using a single NVIDIA V100 GPU. Remarkably, despite BIT not being explicitly pre-trained with contrastive loss, it surpasses prior contrastive learning-based methods, such as CoSP and DrugCLIP, with only a small amount of contrastive fine-tuning.

\subsection{Molecular Property Prediction}
\label{sec:mpp}
In addition to the protein-ligand binding task, we also assess the capabilities of BIT in the molecular property prediction task, where BIT is used as an encoder for small molecules. In this task, we aim to predict the absorption, distribution, metabolism, excretion, and toxicity properties of molecules. We consider eight binary classification datasets from the MoleculeNet benchmark~\citep{wu2018moleculenet}. Following previous studies~\citep{hu2020strategies}, we employ scaffold splitting to divide the dataset into training, validation, and test sets in an 8:1:1 ratio. We use the ROC-AUC as the evaluation metric and report the mean and standard deviation of the results obtained from 3 random seed runs. We compare BIT against representative graph-based pre-trained models, including AttrMask~\citep{hu2020strategies}, ContexPred~\citep{hu2020strategies}, GraphCL~\citep{you2020graph}, InfoGraph~\citep{sun2020infograph}, GROVER~\citep{rong2020self}, MolCLR~\citep{wang2022molecular}, GraphMAE~\citep{hou2022graphmae}, and Mole-BERT~\citep{xia2023molebert}, as well as multimodal pre-trained models, including 3D infoMax~\citep{stark20223d}, GraphMVP~\citep{liu2021pre}, MoleculeSDE~\citep{liu2023group}, and MoleBLEND~\citep{yu2023unified}. The performance of BIT, compared to competitive baselines, is summarized in Table~\ref{tab:moleculenet}. We observe that BIT outperforms the baselines on 6 out of 8 tasks, and achieves an overall relative improvement of 1.9\% in terms of average ROC-AUC compared to the previous state-of-the-art result.

\subsection{Ablation Studies}
We conduct ablation experiments to verify the effectiveness of key design choices in pre-training BIT, and present the results in Table~\ref{tab:ablation}. Based on these results, we observe the following:

\begin{itemize}
\item \textbf{Effect of pre-training data.} Comparing setting [b] with setting [a] reveals the benefits of incorporating small molecule data during pre-training, thereby enhancing the capabilities of BIT as a molecular encoder. When extra pocket data is also included, there is an improvement in performance across all tasks, particularly on binding tasks. Given the limited size of complex data, these findings indicate that pre-training on unbound small molecule and pocket data is effective in acquiring fundamental atom-level knowledge, alleviating the need for bound complex data
\item \textbf{Effect of pre-training tasks.} Eliminating either pre-training objective leads to pronounced declines in performance. We observe that masked token denoising is paramount for 2D representations (see setting [c]), whereas coordinate denoising is indispensable for 3D representations (see setting [d]). These results indicate that our unified pre-training is crucial and yields positive outcomes.
\item \textbf{Effect of MoDE and MoSE} The integration of MoDE and MoSE significantly boosts performance across various tasks (see setting [e]), particularly on the PDBbind dataset, where it is essential to encode both ligands and proteins concurrently while capturing the fine-grained inter-molecular interactions. Such enhancement is in line with our motivation to introduce MoDE and MoSE.
\end{itemize}

\subsection{Real-world virtual screening with BIT}
\label{sec:NMDA}
\begin{figure*}[t]
    \centering
    \includegraphics[width=0.95\linewidth]{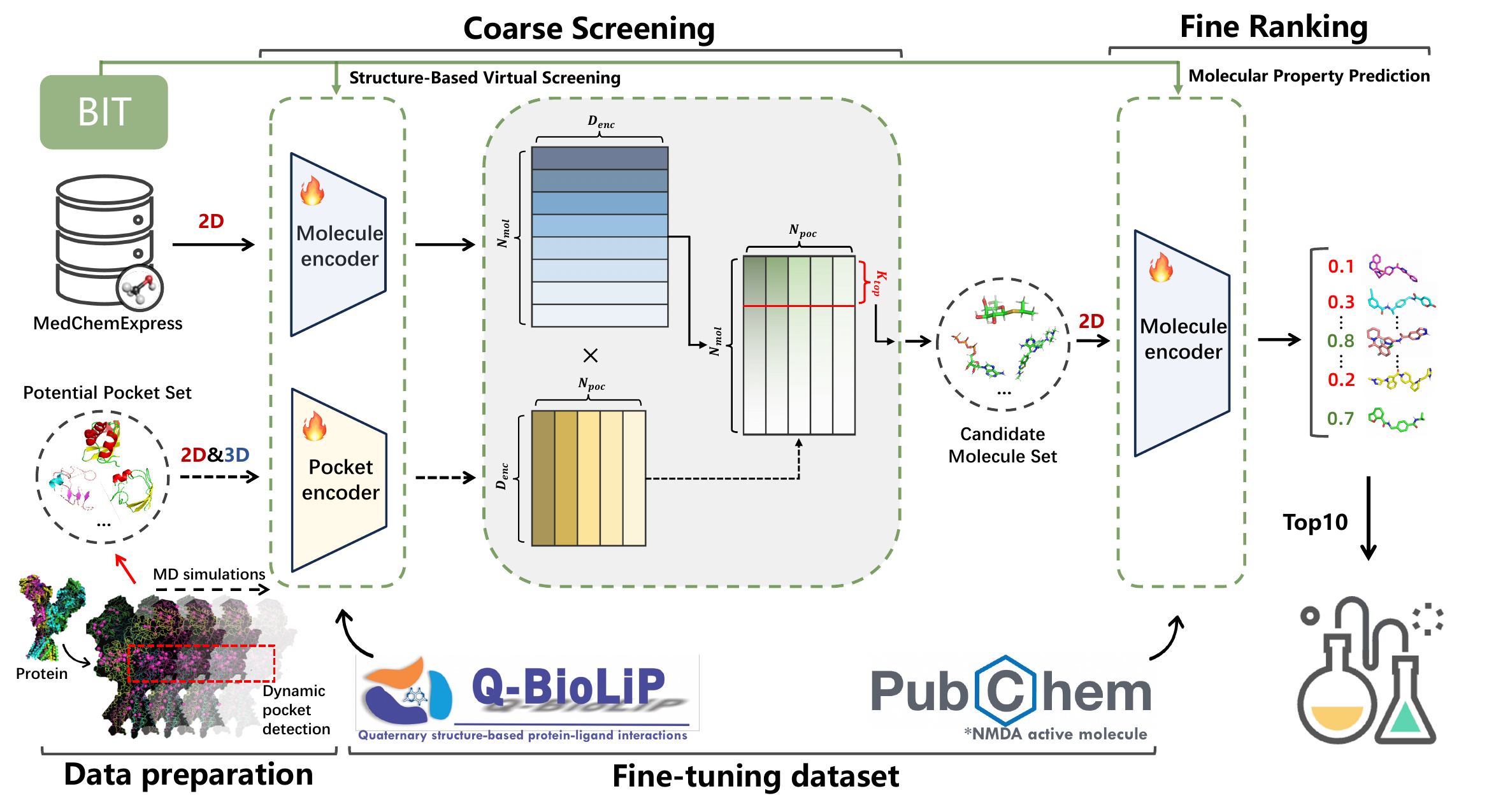}
    \caption{Illustration of the BIT-driven pipeline for virtual screening. BIT serves as a dual encoder for efficient coarse structure-based virtual screening and as a unimodal molecular encoder for fine ligand-based virtual screening.}
    \label{fig:vs_pipeline}
\end{figure*}

We provide an in-depth analysis of BIT's potential to facilitate SBDD across real-world applications. Our goal is to identify new, promising and competitive compounds targeting GluN1/GluN3A N-methyl-D-aspartate (NMDA) receptors~\citep{zhu2020negative, zeng2022identification}. This is achieved through the virtual screening 18 million unique and readily available chemical structures provided by MedChemExpress (MCE). The NMDA receptor is associated with numerous diseases, such as stroke, depression, epilepsy, Alzheimer's disease, and chronic pain, positioning it as a key target for drug development in the treatment of neurological disorders~\citep{vyklicky2014structure,paoletti2013nmda}. The NMDA receptor family is composed of seven subunits: GluN1, GluN2 (2A through 2D), and GluN3 (3A and 3B)~\citep{paoletti2011molecular}. NMDA receptors are heterotetrameric structures that invariably contain at least one GluN1 subunit. The diversity of additional subunits results in various NMDA receptor subtypes, each potentially exhibiting unique functional characteristics~\citep{paoletti2011molecular}. One particular subtype, GluN1/GluN3A, has not been well studied as a therapeutic target due to the lack of small molecule modulators and the absence of crystal structure data. These limitations have hindered further research and complicated drug screening efforts for GluN1/GluN3A~\citep{michalski2024structure}. Consequently, there is a strong need to develop new computational methods to identify potential high-activity molecules that specifically target GluN1/GluN3A receptors, even without available crystal structure information. To address this challenge, we present a coarse-to-fine pipeline driven by BIT that combines structure-based virtual screening and ligand-based virtual screening, as illustrated in~\cref{fig:vs_pipeline}. Below, we outline our strategic approach for efficiently screening an extensive library of drug-like compounds.

\begin{figure*}[t]
    \centering
    \includegraphics[width=0.90\linewidth]{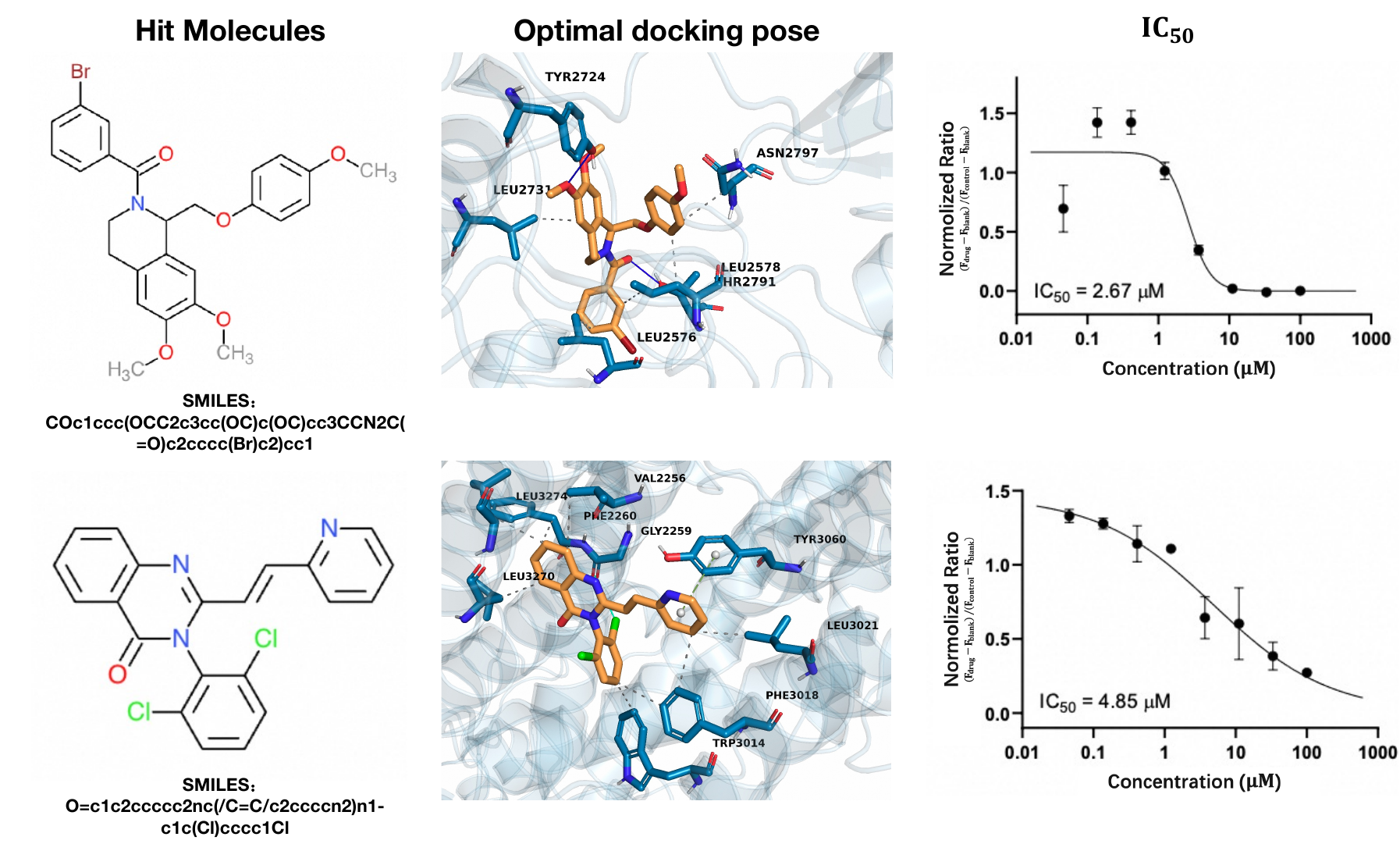}
    \caption{Visualization and experimental validation on identified hit compounds.}
    \label{fig:NMDA}
\end{figure*}

First, we identify potential binding pockets on the GluN1/GluN3A receptor. Due to unavailable crystal structure data for GluN1/GluN3A, we employ homology modeling and molecular dynamics simulations to generate reliable receptor structures. Specifically, we construct the initial structure through homology modeling based on the structure of GluN1/GluN2A~\citep{zhang2021structural}, following the methodology described in~\citet{zeng2022identification}. Subsequently, we perform molecular dynamics simulations using GROMACS~\citep{van2005gromacs} for 900,000 steps, sampling conformations at 100,000-step intervals to obtain 10 distinct structure of the GluN1/GluN3A complex. We adopt P2Rank~\citep{krivak2018p2rank} to identify potential ligand binding sites across all conformations and select the top 100 pockets based on their predicted probability scores.

During the coarse screening stage, BIT functions as a dual encoder for efficient structure-based virtual screening (see~\cref{sec:vs} for details). Specifically, we fine-tune pre-trained BIT on the Q-BioLiP dataset, enhancing its generalization capabilities for virtual screening. We then apply this customized model to screen compounds from three extensive commercial compound libraries provided by MCE: the Bioactive Compound Library Plus, the Commercially Available High-Throughput Screening Library, and MegaUni. Collectively, these three libraries contain 18 million readily available chemical structures. Using fine-tuned BIT, we screen these structures against each detected pocket and ultimately select a total of 300,000 compounds for subsequent analysis. Unlike the coarse screening stage, BIT functions as a unimodal encoder (see~\cref{sec:mpp} for details) during the next fine screening stage, specifically for predicting the probability of binding to the NMDA receptor. We equip BIT with the capability to recognize active molecules targeting NMDA receptors. Given the absence of known active molecules for the GluN1/GluN3A NMDA receptor, we constructed a verified dataset from the publicly available database PubChem~\citep{kim2016pubchem}, consisting of 18,678 samples—12,655 active and 6,023 inactive—related to known NMDA homologous proteins. We then applied BIT, fine-tuned on this dataset, to rank the 300,000 compounds identified in the coarse screening stage. After diversity-based filtering, we ultimately selected 10 candidates for further experimental evaluation.

These candidate compounds underwent an assessment of their biological activity through multi-concentration fluorescence screening, conducted using the FDSS/\textmu Cell high-throughput screening system (Hamamatsu)~\citep{zeng2022identification}. Each compound was prepared in eight different concentrations: 100 \textmu M, 50 \textmu M, 10 \textmu M, 5 \textmu M, 1 \textmu M, 0.5 \textmu M, 0.1 \textmu M, and 0.05 \textmu M. Two of these compounds displayed significant inhibitory effects, $\text{IC}_{50}$ values below 5 \textmu M. In~\cref{fig:NMDA}, we present the experimental validation of the identified active compounds and illustrate the binding mode between the ligands and the protein pockets using AutoDock Vina~\citep{trott2010autodock}. It is noteworthy that the pockets yielding optimal docking results were identified in protein conformations following molecular dynamics simulations, rather than in the initial conformation, underscoring the significance of detecting dynamic pockets. In this scenario, the efficiency of virtual screening becomes particularly crucial due to the increasing number of potential pockets, emphasizing the advantages of BIT over traditional docking software. These two hit molecules exhibit significant potential as starting points for the discovery of new leads and highlight the utility of BIT in advancing SBDD in practical applications.

\section{Discussion}
Molecular representation learning is fundamental to AI-driven drug discovery. Most previous studies learn molecular representations through supervised learning, which constrains their broad applicability in practical scenarios owing to the scarcity of labeled data and suboptimal generalization to out-of-distribution samples. Self-supervised pre-training emerges as a potent solution to these challenges, thanks to the availability of the abundance of unlabeled molecule data: (\expandafter{\romannumeral1}) \textbf{Small molecules}. Initially, researchers employ sequence-based pre-training strategies on string-based molecular data such as SMILES~\citep{wang2019smiles,chithrananda2020chemberta}. As molecular graphs can provide richer 2D topological information, more efforts~\citep{hu2020strategies,rong2020self,wang2022molecular} have focused on pre-training graph neural networks~\citep{xu2018how} or Transformers~\citep{vaswani2017attention} on molecular graphs. Moreover, there are recent studies exploring pre-training on 3D molecular structures to improve performance in predicting molecular properties using geometries~\citep{zaidi2022pre, zhou2023unimol, feng2023fractional}. (\expandafter{\romannumeral2}) \textbf{Proteins}. Protein language models have achieved remarkable success in understanding and generating proteins~\citep{madani2023large, zheng2023structure, zhu2024bridgeif, zhu2024generative} by capturing biological co-evolutionary information from millions of diverse protein sequences~\citep{elnaggar2021prottrans, lin2023evolutionary}, or families of evolutionarily related sequences~\citep{rao2021msa}. Beyond these sequence-based approaches, there is a growing interest in exploring pre-training techniques for protein structures~\citep{zhou2023unimol,zhang2023protein}. While most prior work constructed models based on the characteristics of either small molecules or proteins, our work aims to enhance molecular representation learning by incorporating additional cross-domain relationships learned from biologically relevant protein-ligand complexes.

In this work, we take further strides towards general-purpose molecular modeling. We introduce BIT, a pre-trained foundation model, which is designed to encode molecules across various biochemical domains, including small molecules, proteins, and protein-ligand complexes, in different data formats, including 2D and 3D structures. Experimental results demonstrate that BIT excels across a broad spectrum of protein-ligand binding and molecular learning tasks. Real-world challenges in identifying compounds that bind to the GluN1/GluN3A NMDA receptor further demonstrate the broad applicability and significant potential of the proposed BIT in SBDD.

We compare BIT with related pre-training works to highlight its advantages and unique contributions. Transformer-M~\citep{luo2023one} is a pioneering model capable of processing both 2D and 3D data. However, it lacks a specialized design to capture domain-level specificity, which restricts its transferability between domains. A common workaround is to train separate models for different domains, followed by integrating a simple interaction module, similar to the strategy used by Uni-Mol~\citep{zhou2023unimol}. Yet, this approach is confined to capturing the intra-molecular interactions and inadequately captures the more intricate inter-molecular interactions. In comparison, BIT accommodates domain-specific encoding and cross-domain interactions. Concurrently, DrugCLIP~\citep{gao2023drugclip} employs multimodal learning to align representations of pockets and molecules, facilitating SBDD. Nevertheless, its reliance on contrastive learning limits its ability to capture fine-grained inter-molecular atomic interactions, and it is primarily used for virtual screening. In contrast, BIT excels at discerning fine-grained interactions and is versatile across a wider range of downstream tasks.

BIT's focus on pocket regions enables a nuanced understanding of the protein's active sites, which are crucial for ligand binding. However, one significant limitation of the current BIT is its inability to model the entire protein. As a result, BIT struggles to generalize to downstream tasks that require modeling of the whole protein, such as predicting protein function. Nevertheless, this work serves as a proof of concept for BIT's capacity to model molecular interactions effectively. It is interesting to adopt more efficient attention mechanisms and scale the models to handle entire proteins, thereby extending their applicability to a broader range of tasks.

There are several promising directions for future research: (\expandafter{\romannumeral1}) Investigating a broader array of various, high-quality biomolecules for pre-training could significantly enhance the performance and applicability of our approach. BIT is designed to adapt to any biomolecule and interaction by simply incorporating domain-specific expert networks. (\expandafter{\romannumeral2}) We plan to fine-tune BIT for structure-based molecular generation tasks, such as target protein binding~\citep{luo20213d} and molecular docking~\citep{corso2023diffdock}. (\expandafter{\romannumeral3}) We are working on collecting a more diverse set of real-world and synthetic protein-ligand complexes to support the training of larger models. 

\section{Methods}
\label{sec:method}
\subsection{Input representations}
In biochemical applications, data are collected in the form of molecules represented at different levels of granularity, such as atoms, residues, and nucleobases. However, all molecules can be uniformly represented as sets of atoms held together by attractive or repulsive forces. To more effectively capture and transfer atom-level knowledge across different domains, we propose to share atom embeddings and incorporate domain embeddings to distinguish between small molecules and proteins.
Both small molecule, denoted as $\mathcal{M}$, and protein,  denoted as $\mathcal{P}$, can be represented as a geometric graphs of atoms $\mathcal{G}=(\mathcal{V},\mathcal{E})$. Here $\mathcal{V}=(\mX,\vec{R})$ includes all atoms and $\mathcal{E}$ includes all chemical bonds. In a molecule consisting of $n$ atoms, $\mX \in \mathbb{R}^{n \times d}$ denotes a set of atom feature vectors, $\vec{R} \in \mathbb{R}^{n\times 3}$ denotes a set of atom Cartesian coordinates, and $e_{ij}\in \mathcal{E}$ denotes the feature vector of the edge between atoms $i$ and $j$ if the edge exists. The molecule and protein input representations are computed via summing atom feature embeddings $\mX$, structural positional encodings $\Psi \in \mathbb{R}^{n \times d}$~\citep{ying2021transformers,shi2022benchmarking}, and the corresponding domain-type embedding vectors $\vm_{\text{type}}, \vp_{\text{type}} \in \mathbb{R}^d$. Following~\citet{ying2021transformers}, we introduce special virtual nodes [M\_VNode] for small molecules and [P\_VNode] for proteins, and make connections between the virtual node and each atom node individually.

Given a protein-ligand complex $<\mathcal{M}, \mathcal{P}>$ with cocrystal structures, we identify the binding pocket as the protein atoms located within a minimum distance of 5 {\AA} from the ligand~\citep{muegge1999general}. Then we input the extracted pocket-ligand complex into BIT to learn contextualized representations. It is noteworthy that we only use the binding pocket as the model input rather than the entire protein primarily for the following two reasons: (1) the binding pocket is the paramount region of protein-ligand interaction, experiencing the most significant spatial alterations during the binding process and providing sufficient insight into molecular interactions; (2) the binding pocket contains significantly fewer atoms than the entire protein, leading to lower computational costs and faster training speeds.

\subsection{Backbone}
Recently, several studies have extended the Transformers to model molecules~\citep{rong2020self,ying2021transformers,rampavsek2022recipe,luo2023one}. The vanilla Transformer architecture comprises stacked Transformer blocks~\citep{vaswani2017attention}. Each Transformer block consists of two components: a multi-head self-attention (MSA) layer followed by a feed-forward network (FFN). Layer normalization (LN)~\citep{ba2016layer} is applied after both the MSA and FFN. Let $\mH_{l-1}$ denotes the input, the $l$-th Transformer block works as follows:
\begin{equation}
    \mH_{l}^{\prime} = \mathrm{LN}(\mathrm{MSA}(\mH_{l-1})+\mH_{l-1}), \;\;\;\; \mH_{l} = \mathrm{LN}(\mathrm{FFN}(\mH_{l}^{\prime})+\mH_{l}^{\prime})
\end{equation}
For our general-purpose modeling, we start with Transformer-M~\citep{luo2023one}, a model known for its versatility and effectiveness in handling both 2D or 3D molecule data. Briefly, Transformer-M introduces two separate channels to encode 2D and 3D structural information and integrate them into the MSA module as bias terms. The modified attention matrix $\mA$ is calculated as:
\begin{equation}
    \begin{aligned}
        \mA(\mH) = \mathrm{softmax}\left(\frac{\mH \mW_Q(\mH \mW_K)^{\top}}{\sqrt{d_K}}+ \underbrace{\Phi^{\text{SPD}} + \Phi^{\text{Edge}}}_{\text{2D\ pair-wise\ channel}} + \underbrace{\Phi^{\text{3D Distance}}}_{\text{3D\ pair-wise\ channel}}\right)
        \label{eqn:interaction-encoding}
    \end{aligned}
\end{equation}
where $\mW_Q, \mW_K \in \R^{d \times d_K}$ are learnable weight matrices, the 2D terms ($\Phi^{\text{SPD}}$ and $\Phi^{\text{Edge}}$) and the 3D term ($\Phi^{\text{3D Distance}}$) originate from~\citet{ying2021transformers} and~\citet{shi2022benchmarking}, respectively. To simplify the illustration, we omit the attention head index $h$ and layer index $l$. When molecules are associated with specific 2D or 3D structural information, the corresponding channel will be activated, while the other will be disabled. In combination with the dropout-like 2D-3D joint training strategy~\citep{luo2023one}, where the format of structural information for each data instance is randomly selected, Transformer-M learns to identify chemical knowledge from different data formats and generates meaningful semantic representations for each one.

\bmhead{Mixture-of-Domain-Experts}
To further encode molecules across biochemical domains and learn cross-domain molecular representations enriched with molecular interaction knowledge, we propose to extend Transformer-M with a Mixture-of-Domain-Experts (MoDE) mechanism, employing specialized expert networks for different domains. As shown in~\cref{fig:framework}a, each Transformer block in BIT consists of a shared MSA module and two FFNs, presenting domain experts, namely the molecule expert and the protein expert. In contrast to conventional mixture-of-experts layer~\citep{shazeer2017outrageously,fedus2022switch}, which routes input tokens by a trainable gating network, we directly assign an expert to process each atom token based on its molecule data domain. Sharing the MSA module encourages the model to align protein and ligand, while employing MoDE in place of the FFN encourages the model to capture domain-specific knowledge. The Transformer block of BIT can be abstractly summarized as follows:
\begin{align}
\mH_{l}^{\prime} &= \mathrm{LN}(\mathrm{MSA}\text{-}\mathrm{M}(\mH_{l-1})+\mH_{l-1}) \\
\mH_{l} &= \mathrm{LN}(\mathrm{MoDE}\text{-}\mathrm{FFN}(\mH_{l}^{\prime})+\mH_{l}^{\prime})
\end{align}
where $\mathrm{MSA}\text{-}\mathrm{M}$ denotes the variant of MSA used in Transformer-M.

\bmhead{Mixture-of-Structure-Experts}
The distribution of molecular structures across biochemical domains demonstrates considerable disparity, especially between small molecules and pockets. Consequently, using identical parameters to learn this structural information may introduce potential bias. We further introduce a Mixture-of-Structure-Experts (MoSE) mechanism, which employs specialized pair-wise bias expert networks for different domains. As shown in~\cref{fig:framework}c, we delicately design MoSE based on the observation of 2D and 3D structures across various domains, more detail can be found in~\cref{mose}. For 2D pair-wise bias, we simply use distinct bias experts for different domains. For 3D pair-wise bias, we use one set of parameters to learn intra-molecular distances and another set to learn inter-molecular distances.

Thanks to MoDE and MoSE, BIT decouples the encoding process across different domains. As discussed in~\cref{sec:finetune}, BIT can be fine-tuned to function as either a fusion encoder or a dual encoder, depending on the specific formulation of various downstream protein-ligand binding tasks. 

\subsection{Pre-training BIT}
\label{sec:objectives}
We pre-train BIT on protein-ligand complex data, in addition to unbound small molecule and pocket datasets. We use the Q-BioLiP database~\citep{wei2023biolip2} as the complex corpus. To prevent potential overfitting to a limited portion of the chemical space represented by the Q-BioLiP dataset, we additionally incorporate the PCQM4Mv2 dataset~\citep{nakata2017pubchemqc}, which has been widely used for 3D molecular pre-training~\citep{zaidi2022pre, wang2023automated}, and extract potential pockets on proteins from the Protein Data Bank~\citep{berman2000protein}.

To ensure the scalability of the pre-training process, we employ a unified corrupt-then-denoise objective to pre-train BIT. During pre-training, we randomly corrupt the continuous atom coordinates and the categorical atom types of single-domain molecules (i.e., unbound small molecules and pockets) and ligands from protein-ligand complexes, and guide BIT to restore the original states. 

\subsubsection{Coordinate denoising} 
This task aims to learn meaningful representations that capture the inter-atomic interactions within the molecular structure. Theoretically, this objective can be interpreted as learning an approximate molecular force field from equilibrium structures~\citep{zaidi2022pre}. Thus, we can extend coordinate denoising to protein-ligand complexes, as the experimentally-determined cocrystal structures of the complexes typically represent equilibrium conformations and correspond to local energy minima. To further capture the inter-molecular interactions, we encourage the model to restore the corrupted ligand pose based on the information from both the ligand and pocket.

Formally, let $\vec{R}=\{\vec{r}_1,\vec{r}_2,...,\vec{r}_n\},\vec{r}_i\in\mathbb{R}^3$ denote the binding pose of a bound ligand. We perturb it by adding independent and identically distributed ($i.i.d.$) Gaussian noise to its atomic coordinates $\vec{r}_i$. The resulting noisy atom positions are denoted as $\hat{R}=\{\vec{r}_1+\sigma\vec{\epsilon}_1,\vec{r}_2+\sigma\vec{\epsilon}_2,...,\vec{r}_n+\sigma\vec{\epsilon}_n\}$, where $\vec{\epsilon}_i\sim\mathcal{N}(\vec{0},\mI)$ and $\sigma$ is a hyperparameter controlling the noise scale. The model is trained to predict the noise from the noisy input. The output of the last Transformer block is then fed into an SE(3) equivariant prediction head~\citep{shi2022benchmarking}, driven by the denoising loss $\mathcal{L}_{pos}= \frac{1}{|\mathcal{V}|}{\sum_{i\in{V}} \Vert \hat{\vec{\epsilon}}_{i}} - \vec{\epsilon}_{i} \Vert^2$.

\subsubsection{Masked token denoising} 
This task aims to learn fundamental physicochemical information contained within the molecules or complexes by modeling the dependency between their atoms. This task is similar to the masked language modeling (MLM) task used in BERT~\citep{devlin2018bert} and has achieved remarkable performance in molecular pre-training~\citep{hu2020strategies}. As discussed in~\citet{austin2021structured}, MLM can be interpreted as a categorical denoising process. Given an input molecule, we randomly mask $15\%$ of its atoms and predict each masked atom based on its contextualized representation extracted by BIT. The cross-entropy prediction loss is denoted as $\mathcal{L}_{atom}$.

\subsubsection{Overall pre-training objective} 
During pre-training, we seek to minimize the loss functions of all pre-training tasks simultaneously and reach the overall objective function $\mathcal{L} = \mathcal{L}_{pos} + \lambda \mathcal{L}_{atom}$, where $\lambda$ is the balancing hyper-parameter to control the strength of the masked token denoising task.

\subsection{Fine-tuning BIT on downstream tasks}
\label{sec:finetune}
As illustrated in~\cref{fig:framework}b, since BIT is designed to be a general-purpose pre-trained model, it is straightforward to fine-tune it with task-specific data to adapt to various protein-ligand binding tasks: (\expandafter{\romannumeral1}) \textbf{Protein-ligand binding affinity prediction}. As aforementioned, our model can serve as a fusion encoder to model the molecular interactions between proteins and ligands. Therefore, we extract the final encoding vector from the special token [M\_VNode] as the representation of the protein-ligand complexes and feed it to a task-specific prediction head to make the final prediction. (\expandafter{\romannumeral2}) \textbf{Structure-based virtual screening}. We formulate large-scale virtual screening as a pocket-to-ligand retrieval task. In this task, our model is used as a dual encoder to encode both 3D protein pockets and 2D ligands to vectors of equal length. In fine-tuning, the pre-trained model is further optimized on task-specific data using contrastive learning. During inference, we compute representations of the target pocket and all candidate ligands, and then obtain pocket-to-ligand similarity scores of all possible pocket-ligand pairs using dot products. Hits are identified as ligands that exhibit a high level of similarity to the target pocket. This approach allows for much faster inference speeds than fusion encoder-based methods, which require preliminary molecular docking.

\subsection{Experimental details}
\subsubsection{Pre-training setups}
\label{sec:pretrain}
\bmhead{Datasets}
We pre-train BIT using protein-ligand complex data, in addition to large-scale unbound small molecule and pocket datasets. For \textbf{complex data}, we use the Q-BioLiP database~\citep{wei2023biolip2}, which contains 967,085 biological relevant interactions associated with 3D cocrystal structures as of June 14th, 2023. Q-BioLiP is an updated version of the original BioLiP database~\citep{yang2012biolip}, where protein-ligand interactions are based on the quaternary structure rather than the single-chain monomer structure. This alteration provides higher-quality interactions for analyzing the binding mode. Since our primary focus is on regular ligands, i.e., small molecules, we filter out complexes containing metal ions and DNA/RNA ligands. For \textbf{small molecule data}, we utilize the PCQM4Mv2 dataset~\citep{hu2021ogb}, which has 3.4M organic molecules. These molecules are characterized by their 3D structures at equilibrium, calculated using density functional theory. For \textbf{pocket data}, we apply P2Rank~\citep{krivak2018p2rank} to detect potential ligand binding sites on proteins from the Protein Data Bank~\citep{berman2000protein}, which contains 0.2M proteins with experimentally-determined 3D structures, and collect a dataset of 2M pockets.

\bmhead{Training settings}
Our model adopts the same network configuration as Transformer-M~\citep{luo2023one}. We employ a 12-layer Transformer with a hidden size of 768 and 32 attention heads. We use AdamW optimizer~\citep{loshchilov2018decoupled} with the peak learning rate set to 2e-4, and employ a 12k-step warm-up stage followed by a linear decay scheduler. The total training steps are 200k. Each batch contains 1536 samples, including 512 small molecules, 512 protein pockets, and 512 pocket-ligand complexes. We adopt the 2D-3D joint training strategy proposed in~\citet{luo2023one}. In the coordinate denoising objective, noise scale $\sigma$ is set to 0.2. The balancing hyper-parameter $\lambda$ is set to 0.2. All models are trained on 64 NVIDIA Tesla V100 GPUs for approximately 2 days.

\subsubsection{Protein-ligand binding affinity prediction}
\label{app_exp_pdbbind}
\bmhead{Dataset}
\textbf{PDBbind} dataset is a standard benchmark for assessing the performance of models designed to predict binding affinities. The PDBbind v2016 dataset consists of three subsets: the general set, including 13,283 protein-ligand complexes; the refined set, comprising 4,057 complexes selected from the general set for higher data quality, and the core set, consisting of 285 complexes chosen for the highest data quality. We fine-tune the pre-trained BIT using the refined set and conduct testing with the core set. To prevent data leakage, any data instances present in the core set are removed from the refined set. \textbf{CSAR-HiQ dataset} is an additional benchmark resource, commonly employed as an external dataset to further evaluate the generalization ability of models trained on the PDBbind dataset. We obtain an independent test set consisting of 135 samples from the CSAR-HiQ dataset, excluding any samples that are also present in the PDBbind refined set to prevent overlap~\cite{yan2024multi}.

\bmhead{Baselines}
We compare BIT with five families of methods. Linear Regression (LR), Support Vector Regression (SVR), and RF-Score~\citep{ballester2010machine} are ML-based methods. Pafnucy~\citep{stepniewska2018development} and OnionNet~\citep{zheng2019onionnet} are CNN-based methods. GraphDTA methods~\citep{nguyen2021graphdta} encompass a variety of variants, such as GCN, GAT, GIN, and GAT-GCN. SGCN~\citep{danel2020spatial}, GNN-DTI~\citep{lim2019predicting}, DMPNN~\citep{yang2019analyzing}, MAT~\citep{maziarka2020molecule}, DimeNet~\citep{gasteiger2020directional}, CMPNN~\citep{song2020communicative}, and SIGN~\citep{li2021structure} are GNN-based methods. The recently proposed Transformer-M~\citep{luo2023one} and MBP~\citep{yan2024multi} are pre-training methods. 

\bmhead{Evaluation metrics}
Root Mean Square Error (RMSE), Mean Absolute Error (MAE) and Pearson correlation coefficient (R) are defined as:
\begin{equation}
    RMSE = \sqrt{\frac{1}{|\mathcal{D}|}\sum_{i=1}^{|\mathcal{D}|} (\hat{y}_i-y_i)^2},
\end{equation}
\begin{equation}
    MAE = \frac{1}{|\mathcal{D}|}\sum_{i=1}^{|\mathcal{D}|} |\hat{y}_i-y_i| 
\end{equation}
\begin{equation}
    R = \frac{\sum_{i=1}^{|\mathcal{D}|}(\hat{y}_i-\bar{\hat{y}})(y_i-\bar{y})}{\sqrt{\sum_{i=1}^{|\mathcal{D}|}(\hat{y}_i-\bar{\hat{y}})^2(y_i-\bar{y})^2}}
\end{equation}
$\hat{y}_i$ and $y_i$ respectively represent the predicted and experimental binding affinity of the $i$-th complex in dataset $\mathcal{D}$. The standard deviation (SD) is defined as follows:
\begin{equation}
    SD = \sqrt{\frac{1}{|\mathcal{D}|-1}\sum_{i=1}^{|\mathcal{D}|} [y_i - (a+b \hat{y}_i)]^2}
\end{equation}
where $a$ and $b$ are the intercept and the slope of the regression line, respectively. 

\bmhead{Settings}
We fine-tune the pre-trained BIT on the PDBbind dataset. We use AdamW~\citep{loshchilov2018decoupled} as the optimizer and set its hyperparameter $\epsilon$ to 1e-8 and $(\beta_1,\beta_2)$ to (0.9,0.999). The gradient clip norm is set to 5.0. The peak learning rate is set to 1e-5. The total number of epochs is set to 120. The ratio of the warm-up steps to the total steps is set to 0.06. The batch size is set to 32. The dropout ratios for the input embeddings, attention matrices, and hidden representations are set to 0.0, 0.1, and 0.0 respectively. The weight decay is set to 0.0. 

\subsubsection{Structure-based virtual screening}
\label{app_vs}
\bmhead{Dataset}
The \textbf{DUD-E} dataset comprises 102 targets across different protein families. Each target, on average, is assigned 224 binding compounds and over 10,000 decoys. These decoys are physically similar to the active compounds but differ in terms of their topology. We adopt a four-fold cross-validation strategy and use the same data split approach outlined in GraphCNN~\citep{torng2019graph}. In our data splits, we ensure that no two folds contain targets with greater than 75\% sequence identity. The \textbf{LIT-PCBA} dataset is a much more challenging virtual screening benchmark, proposed to address the biased data problem faced by other benchmarks, e.g., DUD-E. Based on dose-response PubChem bioassays, the LIT-PCBA dataset consists of 15 targets and 7844 experimentally confirmed active and 407,381 inactive compounds.

\bmhead{Baselines}
On the DUD-E dataset, we benchmark BIT against diverse approaches, including docking software Vina~\citep{trott2010autodock}, ML-based methods like RF-Score~\citep{ballester2010machine} and NNScore~\citep{durrant2010nnscore}, DL-based methods such as 3DCNN~\citep{ragoza2017protein}, Graph CNN~\citep{torng2019graph}, DrugVQA~\citep{zheng2020predicting}, and AttentionSiteDTI~\citep{yazdani2022attentionsitedti}, as well as pre-training methods such as CoSP~\citep{gao2023co} and DrugCLIP~\citep{gao2023drugclip}. On the LIT-PCBA data, we choose commercial docking methods such as Surflex~\citep{spitzer2012surflex} and Glide-SP~\citep{Glide}, learning-based methods such as Planet~\citep{zhang2023planet}, Gnina~\citep{mcnutt2021gnina}, DeepDTA~\citep{ozturk2018deepdta}, BigBind~\citep{brocidiacono2023bigbind} and DrugCLIP\citep{gao2023drugclip}.

\bmhead{Evaluation metrics}
Enrichment Factor(EF) is a widely used metric, which is calculated as  
\begin{equation}
    \text{EF}_{\alpha} = \frac{\text{NTB}_{\alpha}}{\text{NTB}_t \times \alpha}\text{,}
\end{equation}
where $\text{NTB}_{\alpha}$ is the number of true binders in the top $\alpha \%$ and  $\text{NTB}_t$ is the total number of binders in the entire screening pool.

ROC enrichment metric (RE) is calculated as a ratio of the true positive rate to the false positive rate (FPR) at a given FPR threshold:
\begin{equation}
   \text{RE}(x\%) = \frac{\text{TP} \times n}{\text{P} \times \text{FP}_{x\%}}\text{,}
\end{equation}
where $n$ is the total number of compounds, $\text{TP}$ is the number of compounds that are correctly identified as active, $\text{P}$ is the total number of active compounds, and $\text{FP}_{x\%}$ is the number of false positives predicted at a specified rate (e.g. 0.5\%, 1\%, etc.).

\bmhead{Settings}
We fine-tune the pre-trained BIT on the DUD-E dataset. We use AdamW~\citep{loshchilov2018decoupled} as the optimizer and set its hyperparameter $\epsilon$ to 1e-8 and $(\beta_1,\beta_2)$ to (0.9,0.999). The gradient clip norm is set to 5.0. The peak learning rate is set to 2e-4. The total number of epochs is set to 10. The ratio of the warm-up steps to the total steps is set to 0.06. The batch size is set to 16. The dropout ratios for the input embeddings, attention matrices, and hidden representations are set to 0.0, 0.1, and 0.0 respectively. The weight decay is set to 0.0. 

\subsubsection{Molecular Property Prediction}
\label{app:moleculenet}
\bmhead{Dataset}
We consider eight binary classification datasets from the MoleculeNet benchmark~\citep{wu2018moleculenet}. Following previous studies~\citep{hu2020strategies}, we employ scaffold splitting to divide the dataset into training, validation, and test sets in an 8:1:1 ratio. The details of the eight datasets used in this work are described below.
\begin{itemize}
    \item BBBP: Blood-brain barrier penetration (BBBP) contains the ability of small molecules to penetrate the blood-brain barrier.
    \item Tox21: The dataset contains toxicity measurements of 8k molecules for 12 targets.
    \item ToxCast: This dataset is derived from toxicology data from in vitro high-throughput screening and contains toxicity measurements for 8k molecules against 617 targets.
    \item SIDER: The Side Effect Resource (SIDER) contains side effects of drugs on 27 system organs. These drugs are not only small molecules but also some peptides with molecular weights over 1000.
    \item ClinTox: This dataset contains the toxicity of the drug in clinical trials and the status of the drug for FDA approval.
    \item MUV: Maximum Unbiased Validation (MUV) is another subset of PubChem BioAssay, containing 90k molecules and 17 bioassays.
    \item HIV: This dataset contains 40k compounds with the ability to inhibit HIV replication.
    \item BACE: This dataset contains the results of small molecules as inhibitors of binding to human $\beta$-secretase 1 (BACE-1).
\end{itemize}

\bmhead{Settings}
We use a grid search to find the best combination of hyperparameters for the molecular property prediction task. The specific search space is shown in Table~\ref{tab:gridsearch}. In all experiments, we choose the checkpoint with the lowest validation loss, and report the results on the test set run by that checkpoint.

\subsubsection{Real-world virtual screening}
\bmhead{Datasets}
We conducted a search for NMDA-related BioAssays in PubChem to select molecules and their corresponding labels. For a subset of unlabeled samples, we established labels by applying $\text{IC}_{50}$ threshold derived from experimental data. This search yielded 6,988 BioAssays, encompassing 18,678 samples, with 12,655 classified as active and 6,013 as inactive based on an $\text{IC}_{50}$ threshold of 10 \textmu mol/L. Molecules below this threshold were deemed active, while those exceeding it were deemed inactive. Samples lacking labels or $\text{IC}_{50}$ values were excluded from the dataset.

\bmhead{Multi-concentration fluorescence screening}
Fluorescence-based screening of the GluN1/GluN3A NMDA receptor was conducted using the FDSS/\textmu Cell high-throughput screening system (Hamamatsu)~\citep{zeng2022identification}. The main objective was to generate dose-response curves for each candidate molecule at multiple concentrations, allowing the determination of the $\text{IC}_{50}$. The following outlines the detailed experimental procedure.

\begin{enumerate}
    \item Experimental Preparation
    \begin{enumerate}
        \item \textbf{Preparation of Candidate Compounds.} The candidate compounds were procured from MedChemExpress (MCE). Each compound was prepared in eight different concentrations: 100 \textmu M, 50 \textmu M, 10 \textmu M, 5 \textmu M, 1 \textmu M, 0.5 \textmu M, 0.1 \textmu M, and 0.05 \textmu M.
        \item \textbf{Cell Line Selection.} The HEK-293 cell line, which stably expresses the NMDA GluN1/GluN3A receptor, was selected for the experiment. Cells were cultured in DMEM media and maintained in a 37°C incubator with $5 \% \text{CO}_2$.
        \item \textbf{Selection of Fluorescent Probe.} The calcium ion fluorescent probe, Fluo-4, was chosen for its high sensitivity in detecting intracellular calcium fluctuations. It allows real-time monitoring in large-scale, automated high-throughput screening experiments.
        \item \textbf{Plate Selection.} 384-well plates were used, with approximately 10,000 cells seeded in each well.
    \end{enumerate}
    \item Experimental Procedure
    \begin{enumerate}
        \item \textbf{Cell Seeding.} The selected HEK-293 cells were seeded into the 384-well plates, ensuring appropriate cell density in each well. After seeding, the plates were incubated at 37°C in a $5 \% \text{CO}_2$ incubator for 24-28 hours until cells reached 80-90\% confluence.
        \item \textbf{Preparation of Compound Solutions.} A series of candidate compound solutions were prepared at concentrations of 100 \textmu M, 50 \textmu M, 10 \textmu M, 5 \textmu M, 1 \textmu M, 0.5 \textmu M, 0.1 \textmu M, and 0.05 \textmu M.
        \item \textbf{Compound Addition.} An automated liquid handling system was used to add the prepared solutions of different concentrations to each well.
        \item \textbf{Fluorescent Probe Addition.} Fluo-4 calcium ion probe was added at a final concentration of 2.5 \textmu M. The plates were incubated for 60 minutes to ensure complete probe entry into the cells and binding with the target molecules.
    \end{enumerate}
    \item Fluorescence Signal Detection
    \begin{enumerate}
        \item \textbf{FDSS/\textmu Cell Setup and Real-Time Monitoring.} The FDSS/\textmu Cell high-throughput screening system was set up with excitation and emission wavelengths at 480 nm and 540 nm, respectively. The system was configured to collect real-time data, acquiring measurements every minute to capture cellular responses following receptor activation.
        \item \textbf{Real-Time Data Collection.} The FDSS/\textmu Cell system automatically monitored and recorded fluorescence intensity data for each well, reflecting the effects of different concentrations of the candidate compounds on NMDA receptor activity.
    \end{enumerate}
    \item Dose-Response Curve Generation
    \begin{enumerate}
        \item \textbf{Using the fluorescence intensity data for each compound at varying concentrations, dose-response curves were generated.} The x-axis represented the compound concentration, while the y-axis displayed the normalized fluorescence intensity. The data was fitted to a four-parameter logistic (4-PL) model to calculate the $\text{EC}_{50}$ or $\text{IC}_{50}$ values for each compound.
    \end{enumerate}
\end{enumerate}

\backmatter

\section*{Declarations}
\subsection*{Data availability}
Datasets used in all benchmark studies have been published previously. The Q-BioLiP dataset can be found at \url{https://yanglab.qd.sdu.edu.cn/Q-BioLiP}. The PCQM4Mv2 dataset can be obtained from \url{http://ogb-data.stanford.edu/data/lsc/pcqm4m-v2-train.sdf.tar.gz}. The Protein Data Bank database can be found at \url{https://www.rcsb.org}. The PDBbind dataset is available at \url{http://www.pdbbind.org.cn}. The CSAR-HiQ dataset can be obtained from \url{http://www.csardock.org}. The DUD-E dataset can be obtained from \url{https://dude.docking.org}. The LIT-PCBA dataset can be obtained from \url{https://drugdesign.unistra.fr/LIT-PCBA}. The MoleculeNet benchmark is available at \url{https://moleculenet.org}.

\newpage

\begin{table*}[t]
\caption{Binding affinity prediction results on the \textit{PDBbind core set} and \textit{CSAR-HiQ set}. We report the official results of baselines from~\citet{li2021structure,luo2023one}. The best results are marked bold.}
\label{tab:affinity}
\centering
\resizebox{\textwidth}{!}{
\begin{tabular}{cl|cccc|cccc}
\toprule
\multicolumn{2}{c|}{\multirow{2}{*}{Method}} & \multicolumn{4}{c|}{PDBbind core set} & \multicolumn{4}{c}{CSAR-HiQ set} \\
\cmidrule{3-10}
    &   & RMSE $\downarrow$  & MAE $\downarrow$  & SD $\downarrow$  & R $\uparrow$	& RMSE $\downarrow$  & MAE $\downarrow$  & SD $\downarrow$  & R $\uparrow$  \\ \midrule 
    \multirow{3}{*}{\shortstack{ML-based \\ Methods}}
    & LR &1.675 (0.000)  &1.358 (0.000)   &1.612 (0.000)  &0.671 (0.000)  &2.071 (0.000)  & 1.622 (0.000)  &1.973 (0.000)  &0.652 (0.000) \\  
    & SVR &1.555 (0.000)  &1.264 (0.000)  &1.493 (0.000)  &0.727 (0.000)  &1.995 (0.000)  &1.553 (0.000)  &1.911 (0.000)  &0.679 (0.000) \\
    & RF-Score~\citep{ballester2010machine} &1.446 (0.008)  &1.161 (0.007)  &1.335 (0.010)  &0.789(0.003)  &1.947 (0.012)  &1.466 (0.009)  &1.796 (0.020)  &0.723 (0.007)
    \\ \midrule 
    
    \multirow{2}{*}{\shortstack{CNN-based \\ Methods}}
    & Pafnucy~\citep{stepniewska2018development} & 1.585 (0.013)  &1.284 (0.021)  &1.563 (0.022)  &0.695 (0.011)  &1.939 (0.103)  &1.562 (0.094)  &1.885 (0.071) &0.686 (0.027) \\  
    & OnionNet~\citep{zheng2019onionnet} &1.407 (0.034)  &1.078 (0.028)  &1.391 (0.038)  &0.768 (0.014)  &1.927 (0.071)  &1.471 (0.031)  &1.877 (0.097)  &0.690 (0.040)
    \\ \midrule 
    
    \multirow{4}{*}{\shortstack{GraphDTA \\ Methods}}
    & GCN &1.735 (0.034)  &1.343 (0.037)  &1.719 (0.027)  &0.613 (0.016)  &2.324 (0.079)  &1.732 (0.065)  &2.302 (0.061)  &0.464 (0.047) \\  
    & GAT &1.765 (0.026)  &1.354 (0.033)  &1.740 (0.027)  &0.601 (0.016)  &2.213 (0.053)  &1.651 (0.061)  &2.215 (0.050)  &0.524 (0.032) \\
    & GIN &1.640 (0.044)  &1.261 (0.044)  &1.621 (0.036)  &0.667 (0.018)  &2.158 (0.074)  &1.624 (0.058)  &2.156 (0.088)  &0.558 (0.047) \\
    & GAT-GCN &1.562 (0.022)  &1.191 (0.016)  &1.558 (0.018)  &0.697 (0.008)  &1.980 (0.055)  &1.493 (0.046)  &1.969 (0.057)  &0.653 (0.026)
    \\ \midrule 
    
    \multirow{8}{*}{\shortstack{GNN-based \\ Methods}}
        & GraphDTA~\citep{nguyen2021graphdta} &1.562 (0.022)  &1.191 (0.016)  &1.558 (0.018)  &0.697 (0.008)  &1.980 (0.055)  &1.493 (0.046)  &1.969 (0.057)  & 0.653 (0.026) \\
    & SGCN~\citep{danel2020spatial} &1.583 (0.033)  &1.250 (0.036)  &1.582 (0.320)  &0.686 (0.015)  &1.902 (0.063)  &1.472 (0.067)  &1.891 (0.077)  &0.686 (0.030) \\  
    & GNN-DTI~\citep{lim2019predicting} &1.492 (0.025)  &1.192 (0.032)  &1.471 (0.051)  &0.736 (0.021)  &1.972 (0.061)  &1.547 (0.058)  &1.834 (0.090)  &0.709 (0.035) \\
    & DMPNN~\citep{yang2019analyzing} &1.493 (0.016)  &1.188 (0.009)  &1.489 (0.014)  &0.729 (0.006)  &1.886 (0.026)  &1.488 (0.054)  &1.865 (0.035)  &0.697 (0.013) \\
    & MAT~\citep{maziarka2020molecule} &1.457 (0.037)  &1.154 (0.037)  &1.445 (0.033)  &0.747 (0.013)  &1.879 (0.065)  &1.435 (0.058)  &1.816 (0.083)  &0.715 (0.030) \\
    & DimeNet~\citep{gasteiger2020directional} &1.453 (0.027)  &1.138 (0.026)  &1.434 (0.023)  &0.752 (0.010)  &1.805 (0.036)  &1.338 (0.026)  &1.798 (0.027)  &0.723 (0.010) \\
    & CMPNN~\citep{song2020communicative} &1.408 (0.028)  &1.117 (0.031)  &1.399 (0.025)  &0.765 (0.009)  &1.839 (0.096)  &1.411 (0.064)  &1.767 (0.103)  &0.730 (0.052) \\ 
    & SIGN~\citep{li2021structure} &1.316 (0.031)  &1.027 (0.025)  &1.312 (0.035)  &0.797 (0.012)  &1.735 (0.031)  &1.327 (0.040)  &1.709 (0.044)  &0.754 (0.014)
    \\ \midrule 

        \multirow{2}{*}{\shortstack{Pre-training \\ Methods}}
        & MBP~\citep{yan2024multi} &1.263 (0.023) &0.999 (0.024) &1.229 (0.026) &0.825 (0.008) &1.624 (0.037) &1.240 (0.038) &1.536 (0.052) &0.791 (0.016) \\
        & Transformer-M~\citep{luo2023one} &1.232 (0.013) &0.940 (0.006) &1.207 (0.007) &0.830 (0.011) & - & - & - & -
        \\ \midrule 

   Ours & BIT & \bf 1.175 (0.010) & \bf 0.919 (0.002) & \bf 1.166 (0.014) & \bf 0.845 (0.004) & \bf 1.522 (0.021) & \bf 1.158 (0.021) & \bf 1.377 (0.026) & \bf 0.838 (0.006) \\ 
\bottomrule 
\end{tabular}
}
\end{table*}

\begin{table}[h]
\caption{Virtual screening results on the DUD-E dataset. We report the official results of baselines from~\citet{yazdani2022attentionsitedti,gao2023drugclip}.}
\label{tab:dud}
\begin{tabular*}{\textwidth}{@{\extracolsep\fill}lrrrrr}
\toprule
Method     & AUC $\uparrow$ & $\text{RE}_{0.5\%}$ $\uparrow$ & $\text{RE}_{1.0\%}$ $\uparrow$ & $\text{RE}_{2.0\%}$ $\uparrow$ & $\text{RE}_{5.0\%}$ $\uparrow$ \\
\midrule
Vina~\citep{trott2010autodock}      & 71.6    & 9.14   & 7.32   & 5.88   & 4.44   \\
\midrule
NNScore~\citep{durrant2010nnscore}   & 58.4    & 4.17   & 2.98   & 2.46   & 1.89   \\
RF-Score~\citep{ballester2010machine}  & 62.2    & 5.63   & 4.27   & 3.50   & 2.68   \\
\midrule
3DCNN~\citep{ragoza2017protein}     & 86.8    & 42.56  & 29.65  & 19.36  & 10.71  \\ 
Graph CNN & 88.6    & 44.41  & 29.75  & 19.41  & 10.74  \\
DrugVQA~\citep{torng2019graph}   & 97.2    & 88.17  & 58.71  & 35.06  & 17.39  \\
AttentionSiteDTI~\cite{yazdani2022attentionsitedti} 
          & 97.1	  & 101.74 & 59.92	& 35.07	 & 16.74  \\
CoSP~\citep{gao2023co}      & 90.1    & 51.05  & 35.98  & 23.68  & 12.21  \\   
DrugCLIP~\citep{gao2023drugclip}  & 96.6    & 118.10 & 67.17  & 37.17  & 16.59  \\
\midrule
BIT      & \bf 97.6 & \bf 147.76 & \bf 78.50 & \bf 41.93 & \bf 17.98  \\
\bottomrule
\end{tabular*}
\end{table}

\begin{table}[h]
\caption{Virtual screening results on the LIT-PCBA dataset.}
\label{tab:lit}
\begin{tabular*}{\textwidth}{@{\extracolsep\fill}lrrrr}
\toprule
Method & AUC $\uparrow$ & $\text{EF}_{0.5\%}$ $\uparrow$ & $\text{EF}_{1.0\%}$ $\uparrow$ & $\text{EF}_{5.0\%}$ $\uparrow$ \\
\midrule
Surflex~\citep{spitzer2012surflex}   & 51.47    & -   & 2.50   & -      \\
Glide-SP~\citep{Glide}  & 53.15    & 3.17   &3.41   & 2.01      \\
\midrule
Planet~\citep{zhang2023planet}      & 57.31    & 4.64   & 3.87   & 2.43      \\
Gnina~\citep{mcnutt2021gnina}     & 60.93    & -  & 4.63  & -    \\ 
DeepDTA~\citep{ozturk2018deepdta} & 56.27    & -  & 1.47  & -    \\
BigBind~\citep{brocidiacono2023bigbind}      & 60.80    & -  & 3.82  & -    \\   
DrugCLIP~\citep{gao2023drugclip}   & 57.17    & 8.56  & 5.51  & 2.27    \\
\midrule
BIT      & \bf 61.04 & \bf 10.02 & \bf 5.76 & \bf 2.67   \\   
\bottomrule
\end{tabular*}
\end{table}

\begin{table*}[h]
\caption{Molecular property prediction results (with 2D topology only) on the MoleculeNet benchmark. The best and second best results are marked \uline{bold} and bold, respectively.}
\label{tab:moleculenet}
\centering
\resizebox{\textwidth}{!}{
\begin{tabular}{lccccccccc}
\toprule
Methods & BBBP $\uparrow$ & Tox21 $\uparrow$ & ToxCast $\uparrow$ & SIDER $\uparrow$ & ClinTox $\uparrow$ & MUV $\uparrow$ & HIV $\uparrow$ & BACE $\uparrow$ & Avg $\uparrow$ \\ \midrule
AttrMask~\citep{hu2020strategies} & 65.0$\pm$2.36 & 74.8$\pm$0.25 & 62.9$\pm$0.11 & 61.2$\pm$0.12 & \textbf{87.7$\pm$1.19} & 73.4$\pm$2.02 & 76.8$\pm$0.53 & 79.7$\pm$0.33 & 72.68 \\
ContextPred~\citep{hu2020strategies} & 65.7$\pm$0.62 & 74.2$\pm$0.06 & 62.5$\pm$0.31 & 62.2$\pm$0.59 & 77.2$\pm$0.88 & 75.3$\pm$1.57 & 77.1$\pm$0.86 & 76.0$\pm$2.08 & 71.28 \\
GraphCL~\citep{you2020graph} & 69.7$\pm$0.67 & 73.9$\pm$0.66 & 62.4$\pm$0.57 & 60.5$\pm$0.88 & 76.0$\pm$2.65 & 69.8$\pm$2.66 & 78.5$\pm$1.22 & 75.4$\pm$1.44 & 70.78 \\
InfoGraph~\citep{sun2020infograph} & 67.5$\pm$0.11 & 73.2$\pm$0.43 & 63.7$\pm$0.50 & 59.9$\pm$0.30 & 76.5$\pm$1.07 & 74.1$\pm$0.74 & 75.1$\pm$0.99 & 77.8$\pm$0.88 & 70.96 \\
GROVER~\citep{rong2020self} & 70.0$\pm$0.10 & 74.3$\pm$0.10 & 65.4$\pm$0.40 & 64.8$\pm$0.60 & 81.2$\pm$3.00 & 67.3$\pm$1.80 & 62.5$\pm$0.90 & 82.6$\pm$0.70 & 71.01 \\
MolCLR~\citep{wang2022molecular} & 66.6$\pm$1.89 & 73.0$\pm$0.16 & 62.9$\pm$0.38 & 57.5$\pm$1.77 & 86.1$\pm$0.95 & 72.5$\pm$2.38 & 76.2$\pm$1.51 & 71.5$\pm$3.17 & 70.79 \\ 
GraphMAE~\citep{hou2022graphmae} & 72.0$\pm$0.60 & 75.5$\pm$0.60 & 64.1$\pm$0.30 & 60.3$\pm$1.10 & 82.3$\pm$1.20 & 76.3$\pm$2.40 & 77.2$\pm$1.00 & 83.1$\pm$0.90 & 73.85 \\
Mole-BERT~\citep{xia2023molebert} & 71.9$\pm$1.60 & 76.8$\pm$0.50 & 64.3$\pm$0.20 & 62.8$\pm$1.10 & 78.9$\pm$3.00 & 78.6$\pm$1.80 & 78.2$\pm$0.80 & 80.8$\pm$1.40 & 74.04\\ \midrule
3D InfoMax~\citep{stark20223d} & 69.1$\pm$1.07 & 74.5$\pm$0.74 & 64.4$\pm$0.88 & 60.6$\pm$0.78 & 79.9$\pm$3.49 & 74.4$\pm$2.45 & 76.1$\pm$1.33 & 79.7$\pm$1.54 & 72.34  \\
GraphMVP~\citep{liu2021pre} & 72.4$\pm$1.60 & 74.4$\pm$0.20 & 63.1$\pm$0.40 & 63.9$\pm$1.20 & 77.5$\pm$4.20 & 75.0$\pm$1.00 & 77.0$\pm$1.20 & 81.2$\pm$0.90 & 73.07 \\ 
MoleculeSDE~\citep{liu2023group} & 71.8$\pm$0.76 & 76.8$\pm$0.34 & 65.0$\pm$0.26 & 60.8$\pm$0.39 & 87.0$\pm$0.53 & \textbf{\uline{80.9$\pm$0.37}} & 78.8$\pm$0.92 & 79.5$\pm$2.17 & 75.07 \\
MoleBLEND~\citep{yu2023unified} & \textbf{73.0$\pm$0.81} & \textbf{77.8$\pm$0.89} & \textbf{66.1$\pm$0.03} & \textbf{\uline{64.9$\pm$0.35}} & 87.6$\pm$0.75 & 77.2$\pm$2.38 & \textbf{79.0$\pm$0.89} & \textbf{83.7$\pm$1.46} & \textbf{76.16} \\ 
\midrule
BIT &  \textbf{\uline{73.9$\pm$0.74}} & \textbf{\uline{78.2$\pm$0.77}} & \textbf{\uline{66.4$\pm$0.29}} & \textbf{64.8$\pm$0.51} & \textbf{\uline{91.9$\pm$1.33}} & \textbf{79.4$\pm$0.80} & \textbf{\uline{80.0$\pm$0.51}} & \textbf{\uline{86.1$\pm$1.35}} & \textbf{\uline{77.59}} \\
\bottomrule
\end{tabular}
}
\end{table*}

\begin{table*}[h]
\caption{
Ablation studies of key design choices in BIT.}
\label{tab:ablation}
\centering
\resizebox{\textwidth}{!}{
\begin{tabular}{cccc|cc|c|cccc}
\toprule
& \multicolumn{3}{c}{\textbf{Pre-Training Data}} &
\multicolumn{2}{c}{\textbf{Pre-Training Tasks}} &
\multicolumn{1}{c}{\textbf{Backbone}} &
\multicolumn{2}{c}{\textbf{Property}} & \multicolumn{2}{c}{\textbf{Binding}} \\
& Complex & Molecule & Pocket & Token & Coordinate & MoDE+MoSE & HIV $\uparrow$ & Tox21 $\uparrow$ & PDBbind (MAE) $\downarrow$ & DUD-E (AUC) $\uparrow$ \\
\midrule
w/o pre-training & \xmark & \xmark & \xmark & \xmark & \xmark & \cmark & 70.9 & 75.1 & 1.114  & 94.9 \\
\midrule
\tblidx{a} & \cmark & \xmark & \xmark & \cmark & \cmark & \cmark & 77.5 & 75.3 & 0.945  & 95.1     \\
\tblidx{b} & \cmark & \cmark & \xmark & \cmark & \cmark & \cmark & 79.2 & 78.0 & 0.928  & 96.5     \\
\tblidx{c} & \cmark & \cmark & \cmark & \cmark & \xmark & \cmark & 78.8 & 78.0 & 0.993  & 96.3     \\
\tblidx{d} & \cmark & \cmark & \cmark & \xmark & \cmark & \cmark & 78.5 & 77.5 & 0.940  & 95.7     \\
\tblidx{e} & \cmark & \cmark & \cmark & \cmark & \cmark & \xmark & 78.2 & 76.8 & 0.968  & 97.3     \\
\midrule
BIT & \cmark & \cmark & \cmark & \cmark & \cmark & \cmark & \bf 80.0 & \bf 78.2 & \bf 0.919 & 97.6 \\
\bottomrule
\end{tabular}
}
\end{table*}

\begin{table}[hb]
  \caption{Search space for the MoleculeNet benchmark.}
  \centering
  \begin{tabular*}{\textwidth}{@{\extracolsep\fill}lc}
    \toprule
    Hyperparameter & Search space \\      
  \midrule
  Learning rate   & [2e-5, 5e-5, 1e-4, 2e-4]   \\
  Batch size      & [32, 64, 128, 256]         \\
  Warmup ratio    & [0, 0.06]                  \\
  \bottomrule
  \end{tabular*}
  \label{tab:gridsearch}
\end{table}

\bigskip

\clearpage
\bibliography{sn-bibliography}

\newpage

\begin{center}
    {\large \textbf{Supplementary Information for \\ A Generalist Cross-Domain Molecular Learning Framework for Structure-Based Drug Discovery}}
\end{center}
\vspace{1em} 

\begin{appendices}

\section{Implementation Details}
\subsection{Prediction Head for Position Output} We use the SE(3) equivariant prediction head proposed in~\cite{shi2022benchmarking}:
\begin{equation}
    \begin{aligned}
        \hat{\vec{\epsilon}}^k_i&=(\sum_{v_j \in V}{a_{ij}\Delta^k_{ij}\mX^{(L)}_j\mW^1_N})\mW^2_N,\quad k=0,1,2
        \label{eqn:noise_pred_layer}
    \end{aligned}
\end{equation}
where $\mX^{(L)}_j$ is the output of the last Transformer block, $a_{ij}$ is the attention score between atom $i$ and $j$ calculated by Eqn.\ref{eqn:interaction-encoding}, $\Delta^k_{ij}$ is the k-th element of the directional vector $\frac{\vec{r}_i - \vec{r}_j}{\lVert \vec{r}_i - \vec{r}_j \rVert}$ between atom $i$ and $j$, and $\mW^1_N\in\mathbb{R}^{d\times d}, \mW^2_N\in\mathbb{R}^{d\times 1}$ are learnable weight matrices. 

\section{Further results}
\label{app_results}

\subsection{Investigation on the impact of MoSE}
\label{mose}
\bmhead{The distribution of molecular structures}
\label{ablation:distribution}

\begin{figure*}[t]
    \centering
    \includegraphics[width=0.9\linewidth]{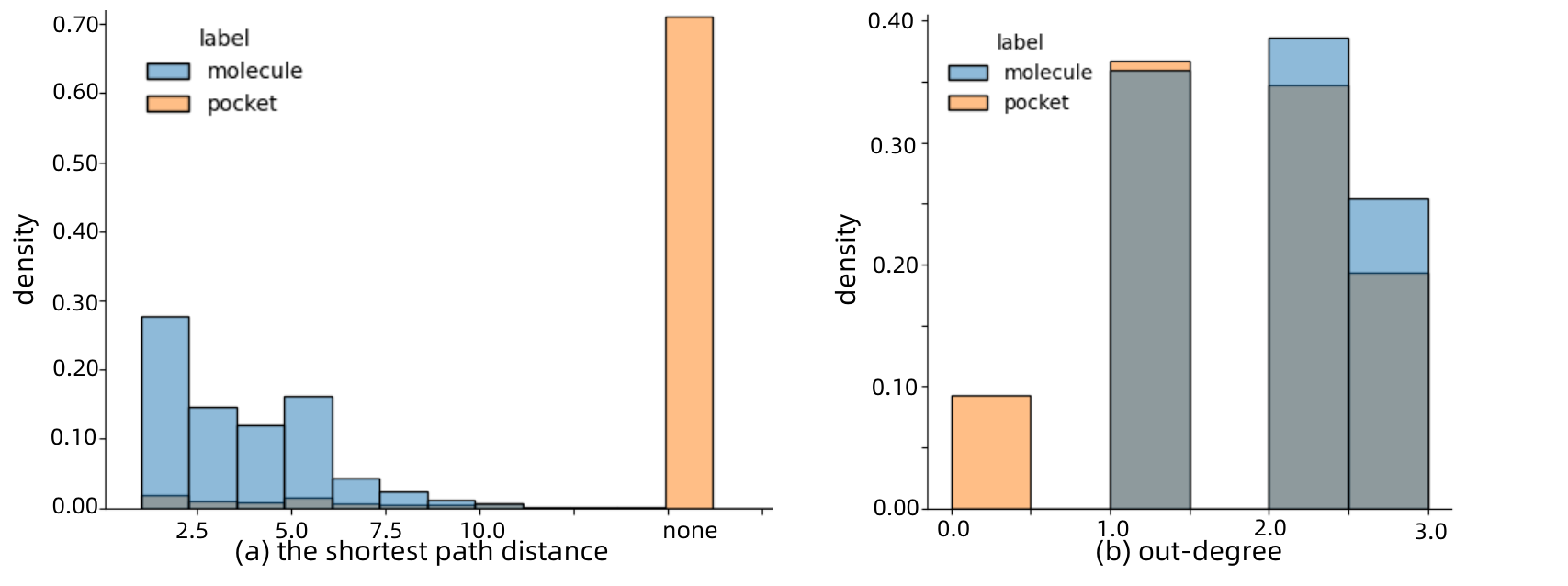}
    \caption{The distribution of 2D structure between molecule and pocket. The gray area indicates the overlapping part. (a) The shorest path distance, none represents that there is no path between the atoms; (b) Out-degree, 0 represents that the atom has no edge connected to other atoms. }
    \label{fig:distribution}
\end{figure*}

In order to further investigate the structural distribution differences between different domains, we randomly sample 10,000 pockets along with small molecules and calculate their the shortest paths of atom-atom pair and out-degree which include 2D structural information and are widely applied in methods involving the addition of bias~\citep{luo2023one, ying2021transformers}. As shown in Figure \ref{fig:distribution}, the results indicate significant differences in the 2D structural information distributions between pockets and small molecules. We think the reason for the difference is that molecules are natural chemical entities, while pockets are artificially extracted from protein sequences by bioinformatics tools or physicochemical rules. 
However, 3D structural information between small molecules and pockets are not significantly different, which can be attributed to their conformations being calculated based on the energy-minimized state of the entity as a biological entity. In other words, their conformations are all derived from interactions based on the same physical and chemical force field. The design methodology of our model is to build the part-specific separately; hence, MoSE has different constructions for 2D and 3D structures.

\bmhead{Ablation Study of MoDE-MoSE Module}
\label{ablation:mose-m}
As presented in Table \ref{tab:ablation-mose}, the integration of MoDE-MoSE significantly boosts performance across various tasks. Results show that MoSE performs better on binding tasks. we think that MoSE could capture the fine-grained inter-molecular interactions.

\bmhead{Ablation Study of MoSE Component}
\label{ablation:mose-c}
we conduct ablation study on the components of MoSE during pre-training. The experimental setup involved training five different models: without MoSE, only 2D-MoSE, only 3D-MoSE* (designed consistent with 2D-MoSE), only 3D-MoSE, and the 2D3D-MoSE. 
Each model is validated on the same validation set.The results are illustrated in Table \ref{tab:ablation-mose-c}. We find that adding 2D-MoSE could lead to a significant improvement, while 3D-MoSE* did not result in a noticeable enhancement, which is consistent with our observations in~\cref{ablation:distribution}. After modifying 3D-MoSE* to 3D-MoSE, we achieved improvement in Complex dataset, this indicates that 3D-MoSE achieves better performance on data containing interactions. Finally, by combining 2D and 3D, we achieve the best performance.

\begin{table}[t]
\caption{
Ablation studies of MoDE/MoSE in BIT.}
\label{tab:ablation-mose}
\centering
\begin{tabular}{c|cc|ccc}
\toprule
&\multicolumn{2}{c}{\textbf{Backbone}} &
\multicolumn{2}{c}{\textbf{Property}} & \multicolumn{1}{c}{\textbf{Binding}} \\
& MoDE & MoSE & HIV $\uparrow$ & Tox21 $\uparrow$ & PDBbind (MAE) $\downarrow$ \\
\midrule
\tblidx{a} & \xmark & \xmark & 78.2 & 76.8 & 0.968      \\
\tblidx{b} & \cmark & \xmark & 78.8 & 77.7 & 0.942      \\
\tblidx{c} & \xmark & \cmark & 78.6 & 77.4 & 0.931      \\
\midrule
BIT & \cmark & \cmark & \bf 80.0 & \bf 78.2 & \bf 0.919 \\
\bottomrule
\end{tabular}
\end{table}

\begin{table}[t]
\caption{
Ablation studies of MoSE component in BIT.}
\label{tab:ablation-mose-c}
\centering
\begin{tabular}{c|cc|cccccc}
\toprule
&\multicolumn{2}{c}{\textbf{Backbone}} &
\multicolumn{2}{c}{\textbf{Complex}} & \multicolumn{2}{c}{\textbf{Ligand}} & \multicolumn{2}{c}{\textbf{Pocket}}\\
& 2D & 3D  &$\mathcal{L}_{pos}$ &$\mathcal{L}$ &$\mathcal{L}_{pos}$ &$\mathcal{L}$ &$\mathcal{L}_{pos}$ &$\mathcal{L}$\\
\midrule
\tblidx{a} & \xmark & \xmark & 0.217 & 0.228 &  0.224     &0.248 &0.289 & 0.295\\
\tblidx{b} & \cmark & \xmark & 0.205 & 0.216 & 0.207      &0.230  &0.284 & 0.290\\
\tblidx{c} & \xmark & \cmark* & 0.220 & 0.231 & 0.219      &0.242 &0.289 &0.294\\
\tblidx{d} & \xmark & \cmark & 0.209 & 0.220 & 0.224      &0.248 &0.286 &0.291 \\
\midrule
BIT & \cmark & \cmark & \bf 0.201 & \bf 0.211 & \bf0.206 & \bf0.229 & \bf0.279 & \bf0.283 \\
\bottomrule
\end{tabular}
\end{table}

\section{More related work}
\label{app_related_work}

\bmhead{Multimodal representation learning} Multimodal representation learning has been extensively studied to enhance understanding across various areas, including image analysis~\citep{radford2021learning}, video processing~\citep{sun2019videobert}, and speech recognition~\citep{ao2022speecht5}. Among these applications, Transformer~\citep{vaswani2017attention, dosovitskiy2021an} has become a critical building block, owing to its flexibility in aligning and integrating information across multimodal data sources. There are three main types of architectures to cater to different multimodal learning requirements: dual encoder~\citep{radford2021learning, jia2021scaling} for efficient retrieval, fusion encoder~\citep{kim2021vilt,li2021align} for deep understanding, and encoder-decoder architectures~\citep{wang2022simvlm} for conditional generation. Some research~\citep{li2022blip, bao2022vlmo, wang2023image} have explored effective ways to integrate the strengths of these architectures. Recently, multimodal learning has also found applications in the biomedical field. There have been early attempts to enhance molecular representation learning by leveraging the correspondence and consistency between 2D topological structures and 3D geometric views~\citep{liu2022multi,stark20223d,liu2023group} or incorporating biomedical text~\citep{liu2022multi, xu2023protst,yin2024multi}.




\end{appendices}

\end{document}